# Fuzzy Knowledge-Based Architecture for Learning and Interaction in Social Robots

**Mehdi Ghayoumi, Maryam Pourebadi**

University of San Diego, University of California San Diego
mghayoumi@sandiego.edu, pourebadi@ucsd.edu

**Abstract**

In this paper, we introduce an extension of our presented cognitive-based emotion model [27][28]and [30], where we enhance our knowledge-based emotion unit of the architecture by embedding a fuzzy rule-based system to it. The model utilizes the cognitive parameters dependency and their corresponding weights to regulate the robot's behavior and fuse their behavior data to achieve the final decision in their interaction with the environment. Using this fuzzy system, our previous model can simulate linguistic parameters for better controlling and generating understandable and flexible behaviors in the robots. We implement our model on an assistive healthcare robot, named Robot Nurse Assistant (RNA) and test it with human subjects. Our model records all the emotion states and essential information based on its predefined rules and learning system. Our results show that our robot interacts with patients in a reasonable, faithful way in special conditions which are defined by rules. This work has the potential to provide better on-demand service for clinical experts to monitor the patients' emotion states and help them make better decisions accordingly.

## Introduction

There is a substantial daily increase in demands to make practical and efficient use of social robotics, especially for educational purposes in the healthcare industry [1, 2,31]. In the social robotics domain, several key factors are affecting to show, what acceptable robot communication is for humans [3, 4, 5, 6]. Recent researches show the interaction between robot and human can be understood, designed, and evaluated basically through three major modalities: speech, gestures, and facial expressions [26][32] and [35]. Moreover, there are many theories on how emotion and cognition are related. James–Lange's work shows that emotional changes arise through the activation of neurons [12, 13, 19]. Also, based on the Cannon-Bard research, emotion and cognition are separate but simultaneous [14] and in this paper, we use this concept to provide robots with a deep understanding of human emotion using environmental data and rules for improving the cognitive aspects of the robot.

Although designing cognitive-based emotion models is a research area which has encouraged many researchers for a long time [5], there is a gap in the literature which needs to be filled by conducting further research on fusing a cognitive-emotion model with a rule-based model to generate this relation and create an acceptable robot's behavior.

The cognition in the robotics deals with different phenomena's, including perception, robot behavioral, cognitive appraisal, and emotion appraisal. However, these units in robots' construction require an additional unit to store, model and deploy all the necessary data and information by adding rules.

In this article, we introduced the concept of a Fuzzy Knowledge-Based System (FKBS) model for generalizing cognitive-emotional systems on different robots that can be customized for various environments. Using the rule-based system is an extension of the model that results in greater flexibility and control, as well as more realistic emotions.

The proposed model is composed of five major units with interconnections that result in an optimized robot behavior over time: Perception, Robot Behavioral, Cognitive Appraisal, Emotion Appraisal, and Fuzzy Knowledge-Based System. We embedded our model in a physical robot to help as a healthcare assistant named Robot Nurse Assistant (RNA). RNA aims to help clinicians promote healthcare by exploring the usage of our advanced cognitive-emotional model within experimental hospital circumstances to oversee clinical consulting services, provide patient care, and maintain enhanced relations and communication between patients and experts.

The rest of the paper is as follows: Section 2 presents the related work. Section 3 describes the theory and details of our methodology. Section 4 shows the experimental results and the last section conclude paper and presents future work.

## Related Work

There are different models of cognition in robots [20]. Some researchers assumed that all robot's actions should help the human to do the right things [21] and, they should have this ability to understand human actions. On the other hand, they should interact with the environment in the proper operation and behavior [22]. For example, Schachter–Singer indicates that general physiological arousal can cause emotional stimulus, which is then interpreted by the brain and expressed to complete emotional experience [18]. Also, Lazarus argues Cognitive Appraisal (CA) is based on emotion, as humans make an unconscious evaluation before expressing any emotion [11].

The cognitive architecture of the robot can be designed based on different applications of the system [6]. Several of the more well-known architectures include: Adaptive Control of Thought-Rational(ACT-R) [4], Connectionist Learning with Adaptive Rule Induction Online (CLARION) [12], Attentive Self-Modifying (ASMO) [8], State, Operator, and Result (SOAR) [6] ICARUS [14], Bratman's philosophical action theory [15], Belief, Desire, Intention (BDI) , [16] and CAIO [17].

One significant part of presenting more realistic interaction with the human is emotion [23], which has a direct connection with cognition [24]. The point for this purpose is how we can have a system which keeps all rules such as the rule that all human actions are right, and how the human behavior can have effect on the robot cognition and behavior that makes the robot's activities as close to human as possible [25]. The model presented here, separates the rule base database as a different part of the knowledge-based system which includes all rules and makes cognition, as the center of any decision in a robot for any action and behavior.

On the other hand, a fuzzy system inside the knowledge base system helps to regulate the rules and input data from perception and robot behavior components using weights and mathematical functions which make the robot's control more flexible and its behavior more realistic.

## Theory and Details of Model

There are several definitions for emotion and cognition from different fields such as psychology, neuroscience to computer science, but most of the time, when researchers try to explain the cognition aspect of the mind, then they include directly or indirectly emotion in their definition [14]. After looking more at the models and theories about the cognition of the mind and emotion, we inferred that for analyzing the emotion, the most recent promising approaches can be cognition oriented. However, there are some models and theories which they don't mention any type of connections between emotion and cognition, recent research shows that with the cognition, emotion phenomena could be described better [15].

Figure 1 shows the cognitive-emotional model consists of these major units. Using the different parts of the perception unit, the system makes its judgments of the real world based on the way it perceives, and then it makes information. The environmental model tracks transactional data between environment and user and describes the characteristics of the environment. The user's emotion and action are two parts of this unit that are designed to be connected to the environment to perceive the user's behaviors. Also, the robot behavioral unit is employed to define the robot's actions and emotion expression. Three remaining units, i.e., Knowledge-Based System (KBS), Emotion Appraisal (EA), and Cognitive Appraisal (CA) are the core of robot behavior controls that are responsible for managing the communications between all parts of the proposed cognitive-based model [7, 8, 9].

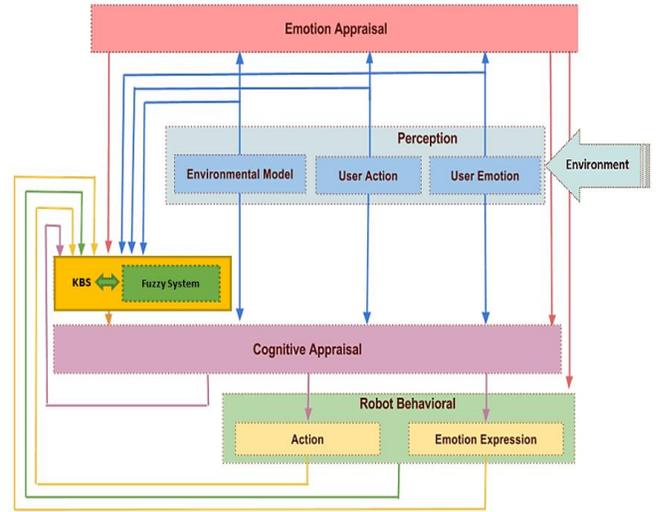

**Figure 1.** The presented architecture has five major units: the input data which come from the environment, Perception, Knowledge-Based System (KBS), Cognitive Appraisal (CA), Emotion Appraisal (EA), and Robot Behavioral (RB). All these parts will describe briefly in the next sections.

### A. Perception

There are three parts in the perception unit, which are connected to the environment to take in the user's behavior. The first part is user emotion states, which can be recognized as either primary or secondary(human), the second part is user action, which is the current actions of the user (human), such as resting, walking, or speaking, and the last part is the environmental model that consists of the environmental parameters which are a set of variables that can affect the robot behavior. The first and second parts are user-oriented, and the third one is the environment-oriented. All decisions in the robot behavior control will start and complete based on the data collected in this section, and then the data will transfer to the KBS, CA, and EA.

## B. Cognitive Appraisal

There are several different architectures for cognition [5]. Some of these architectures such as SOAR, ACTR, and CLARION, are well known and have different techniques. The cognitive appraisal model presented here is designed based on the CLARION architecture, which involves most conditions and helps to generalize the model. The inputs of the cognitive units come from three parts, i.e., perception (in the environmental model, beliefs and goals have the most effects), KBS, and EA. The cognition can directly lead to some actions and emotional states and has direct access to the information from the EA as well. On the other hand, some of the actions and emotion states need to be led indirectly which can be done by different weighted data presented to the CA through the KBS output [16], [17].

The cognitive appraisal is the core of the system in this model, and all processed and unprocessed data are feeding to this part for a final decision (Figure 2). Here, the input data will be initialized and weighted at the beginning, and it will be modified based on the application's demands [33]:

$$C_o = ((W_{EA} \times X_{EA}) + (W_{FKBS} \times X_{FKBS}) + (W_P \times X_P)) \quad (1)$$

Where $(W_{EA} + W_{FKBS} + W_P) = 1$ and Xi are the inputs from each part that can be adapted by the rules in the CA segment.

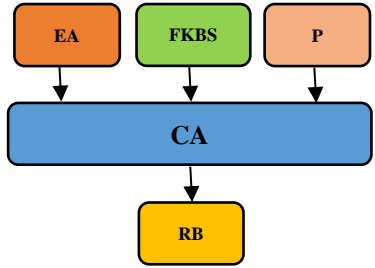

**Figure 2.** The weighted approach in the CA

We assigned the .5 to WFKBS and the .25 to the WEA and WP. We made the WFKBS weight more to make its more effective on robot's behavior. These values can be changed by the system architecture, based on the data types, system needs, and the user's demands.

## C. Emotion Appraisal

According to the Arnold and Lazarus theory, EA is the personal interpretation, evaluation, and explanation of any situation and event, which can cause the emotional state in the robot. Here, in the model, the emotion appraisal section gets the data and information from the perception segment and is communicating with the cognition and KBS directly. The behavior and the emotion appraisal model in the RNA project are based on this architecture [11].

## D. Robot Behavior

The model output is the robot behavior which has two major parts: 1) the robot actions, which are the actions that robot is currently doing or is expected to do such as cooking, driving, or any other skills, and, 2) emotion expression that can be expressed by the facial expressions, gestures, and speech. Both the robot cognition and emotion can change the robot's behavior by direct commands as well as the information regarding the robot's behavior in each step by transferring the data to the KBS. In the RNA project, the actions and emotion expressions are limited to robot-patient interactions and expressions which the clinical expert has the control to customize these based on the patient's needs.

## E. Knowledge-Based System (KBS)

The KBS stores necessary data (can be defined by the system architecture based on the user's demand or application needs) in the memory to make it possible for the system to generate a decision based on the rules, data, and facts. KBS contains a data behavior control mechanism that involves the current robot behavior rules (actions & emotions) in the system, which it can change. It is made up of one inference engine and three types of data (Figure 3): the first part is system facts which some of them are fixed, because they are necessary for all social robots and some may vary based on the robot environment or the application, the second part, is user inference that includes: perception, robot behavioral, cognitive and emotion appraisal, and, the last one is formed from knowledge rules that can be initialized at the beginning of the implementation process but can change over time through applying minor or significant changes by the designer based on the user feedback or expectation from cognition, behavior, and emotion parts. These three parts are connected to the inference engine which is responsible for searching and selecting the rules based on the system knowledge. Finally, the output of the inference engine feeds to the cognitive unit which is going to be used as the main factor to determine the robot behavior.

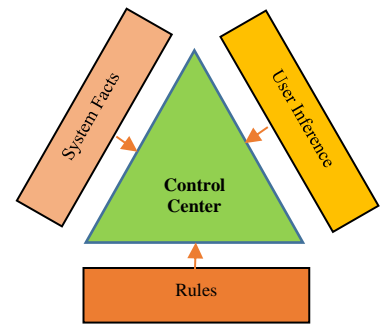

**Figure 3.** The KBS and its parts

The proposed fuzzy knowledge-based system here is simplified in the first step to test the system. More complicated rules based on the robot application and user demand can be added to the system. Also, more parameters such as very low or very high for speech and head rotation values can increase the number of combinations of states for controlling the robot in more specific conditions. For enriching the fuzzy system, some data can be extracted such as skin color which can be added to the actions and rules for better interaction. Also, the current robot behavior and actions are stored in the memory to be used by the designer for future modification and including in the rule-based extension).

Here, the fuzzy knowledge-based system is implemented using the MATLAB toolbox. The data will feed to the fuzzy model, and the outputs will use for robot control. The fuzzy system which deploys here for this implementation has three main parts: Fuzzification, Fuzzy Rule base, and Defuzzification.

### 1) Fuzzification

In the fuzzification step, the real values change into fuzzy values by applying different fuzzifiers (membership functions (MFs)). The data values are classified into three groups: negative, neutral and positive. For a specific value such as x, (for example, $x_n$) the y value for each MFs can be calculated, (this value is between 0 and 1). Any value of negative, neutral and positive emotion states belongs to at least one of the MFs with a certain degree of membership. Also, for speech, we assume three categories based on the sound amplitude: low, normal and high and for the gesture, the system is looking for the head movements in the vertical axis. The head rotation in the vertical axis has 3 classes: normal with 0 degree, 25 for low and high with 45 degree.

### 2) Fuzzy Rule Base

The fuzzy rule base is a part of KBS with this format: "If A then B" where A and B are the rule statements that can be defined by the system designer. In this step, the user can add the desire rules which they can modify the robot control, as well as adapting to the user's feedback or expectations over time. Here, we just focus on some fundamental rules and use them for testing the model. In this phase, three actions (do no action, call the nurses and record data) and three emotion state for recognition (negative: {sadness, fear, disgust, surprise and anger}, neutral and positive: {happiness}) and two emotion states (Neutral and happiness) for expression. for robot behavior and expression have been proposed. We defined and deployed nine rules in this phase as follows:

1) If the state is negative, then do no action and call the nurses and record the data,
2) If the state is neutral, then record the data,
3) If the state is positive, then smile and record the data,
4) If the sound is low and the head angle ratio is low, then do no action and call the nurses and record the data,
5) If the sound is normal, then record the data,
6) If the sound is high and the emotion state is negative, do no action and call the nurses and record the data,
7) If the head angle ratio is low, then do no action and call the nurses and record the data,
8) If the head angle ratio is normal, then record the data,
9) If the head angle ratio is high and the sound is low, then do no action and call the nurses and record the data.

There are also several combinations with some logic operations such as AND and OR which can cover more situations and conditions in which they make the robot's behavior more robust and flexible.

### 3) Defuzzification

In this step, the fuzzy values will be converted to the real values to make a real action for robot behavior. The defuzzification input is a combined output fuzzy set, and its output is a single value. In the fuzzy system, we used the weighted center of gravity for defuzzification. The weighted center formulas and more explanations about it can be found in [24, 25].

## Experimental Results

To test our general architecture for human-robot interaction presented in [29], we used user's facial expression, and partially speech and gesture data for robot action generation and tested it on the Double telepresence robot. To test the extended architecture, we added a fuzzy knowledge-based system to the model and ran the same experiment to test the system. We first trained an identity recognition system with patients and clinicians' facial imagery. This system lets RNA to detect and recognize the known faces, autonomously get close to the person and stand in front of the position with 50-70 centimeters distance. At this point, RNA starts to record the video and sound data and sends it to the server-side. The system receives the data and reports the average of the recognition for different input.

### A. Methodology and Implementation

We are using Double 2 (a telepresence robot), a PC, core i7 with NVIDIA GeForce GTX 1070 on the server-side, and CK+ database for training purposes. The double robot has an iPad on it which has limitations for memory and processing on board, so we capture live camera frames, then feed and processed the frames with CoreImage using the CIDetector developed by Apple, that it easily detects the face. Then the data will be transferred to the PC server for more processing, and finally, the final commands based on the architecture and user inputs are sent to the robot for the proper emotion and action generation. On the server-side, Keras, Python 3.7 and the MATLAB toolbox have been used for creating KBS, training, evaluating, testing, analyzing and processing the data. However, many modifications were made to make the model smaller and more accessible to low-resource environments. Before feeding data to the network, we are doing some data augmentation for ensuring that network is generalized to other variation make adjusting the image brightness or the hue. Resizing and make the database image with the same size for preventing this type of overfitting. Here the proposed CNN model was trained using the CK+ dataset and the frames from the video in the labs from 5 subjects for six different emotion states. In the first convolutional layer, the pre-proceed images will be convolved with 5*5 window, (in total there are 32 filters which it means there are 32 different patterns). The pattern's weights which belong to each pattern will be adjusted as the model learn. The resized image size is 128*128, and they will be 124*124 after the convolution filter, and for 32 filters the layer size is 124*124*32. In the max pooling, there are subsampling outputs with 2*2 size and a stride of 2, and then the output images have 64*64 dimension, and the layer size is 64*64*32. The last pooling layer is used with a softmax activation function and in this layer make emotion classification [34]. To prevent overfitting, we regulated weights by adding a small positive regularization term to the cost function and using dropout just before the last fully connected layer with .5 value. We used the trained model for testing with the image frames which have been sent by a Double robot.

## B. Results

In the first phase of this research, the model was implemented with a few parameters and rules as we discussed in the previous sections and it was tested on a limited virtual condition. Although the robot control and interaction were acceptable, which had reasonable real-time communication with patients, followed the rules and sent the reports to the experts. The cognition part also gives some weights based on the structure that we discussed before. It also can be initialized by the system designer, and its weight also can be adapted to the conditions, the user's feedback and expectations autonomously over time.

Table.1 demonstrates the average accuracy for facial expression. The overall performance is 58.3%, and the most recognized emotions are anger (66.7%) and surprise (63.3%). One of the reasons for misclassification is the same valence-arousal quadrant for example: fear and surprise (12.2%), and sadness as disgust (16.7%).

**Table 1.** Facial Expression accuracy

|        | Ang. | Hap. | Sad  | Surp. | Disg. | Fear |
|--------|------|------|------|-------|-------|------|
| Ang.   | **66.7** | 3.3  | 3.3  | 10    | 6.7   | 3.3  |
| Hap.   | 10   | **50** | 13.3 | 10    | 0     | 13.3 |
| Sad    | 6.7  | 3.3  | **60** | 6.7   | 6.7   | 16.7 |
| Surp.  | 10   | 6.7  | 10   | **63.5** | 10    | 0    |
| Disg.  | 6.7  | 3.3  | 16.7 | 6.7   | **43.2** | 0    |
| Fear   | 3.3  | 6.7  | 0    | 12.2  | 11.1  | **53.3** |

Table.2 demonstrates the average accuracy for speech. The Overall performance is 67%, and the max accuracy is Anger 93%, fear 76%, sadness: 70% Surprise: 60%, disgust: 53% and Happiness; 23.3%. Anger, fear, and sadness had high accuracy because of the energy and speech modulation Happiness obtained a very low recognition rate and was mainly confused with surprise.

**Table 2.** Speech accuracy.

|       | Ang. | Hap. | Sad | Surp. | Disg. | Fear |
|-------|------|------|-----|-------|-------|------|
| Ang.  | **93.3** | 0    | 3.3 | 0     | 3.3   | 0    |
| Hap.  | 10   | **23.3** | 0   | 23.3  | 3.3   | 16.3 |
| Sad   | 16   | 0    | **70** | 10    | 0     | 3.3  |
| Surp  | 13.3 | 10   | 0   | **60** | 13.3  | 0    |
| Disg. | 20   | 10   | 16.7| 13.3  | **53** | 0    |
| Fear  | 0    | 3.3  | 0   | 6.3   | 10    | **76** |

Table.3 demonstrates the average accuracy for the gesture. The accuracies are for Anger: 98.3%, and Happiness: 95.4%, Sadness: 92.1%, surprise: 87.5%, Disgust: 90.2%, fear: 83.3% and misclassifications are Anger is misclassified as disgust (4.3%) and Mix-up of emotions during transition are: {anger, disgust}, {happiness, surprise} and {sadness, disgust, fear}.

**Table 3.** Gesture accuracy

|        | Ang. | Hap. | Sad | Surpr. | Disg. | Fear |
|--------|------|------|-----|--------|-------|------|
| Ang.   | **98.3** | 0    | 0   | 0      | 4.3   | 0    |
| Hap.   | 0    | **95.4** | 0   | 7.2    | 0     | 0    |
| Sad    | 3.1  | 0    | **92.1** | 0      | 2.7   | 2.2  |
| Surpr. | 9.3  | 10.4 | 0   | **87.5** | 2.3   | 5.8  |
| Disg.  | 7.2  | 0    | 4.1 | 3.3    | **90.2** | 4.2  |
| Fear   | 0    | 0    | 0   | 12.2   | 11.1  | **83.3** |

We also gave the subjects a survey which includes three different questions and asked them to rate the robot functionalities. The rate criteria are as follow poor=1, good=2, and very good =3. Table 4 shows the subject's feedback about robot performance.

**Table 4.** subject's feedback

|         | User-Friendly | Real-time response | Proper results |
|---------|---------------|--------------------|----------------|
| $S_1$   | 2             | 2                  | 3              |
| $S_2$   | 2             | 1                  | 3              |
| $S_3$   | 2             | 1                  | 3              |
| $S_4$   | 3             | 2                  | 3              |
| $S_5$   | 2             | 1                  | 2              |
| Average | **2.2**       | **1.4**            | **2.8**        |

## Conclusion and Future Works

In this paper, we presented a fuzzy knowledge-based model for social robots consisting of the rule-based system and cognitive appraisal units. The proposed robot is an assistive social robot: "Robot Nursing Assistant (RNA)" that helps clinicians to have more interaction with patients, follow up with self-reported symptoms, and prepare related reports. Using the system facts and the five major units gives the designer an opportunity to change the robot behavior based on the rules in the KBS. The model can be specified and analyzed based on different environmental models and user's requests. The methodology described has been through preliminary testing with rules on the double robot.

The next step in our research is to enrich the rule database with expert consults and target specific health care operations and then test with patients and experts and evaluate the model based upon on user's feedback for its realistic aspect and the degree of acceptability in interaction with people and their expectations.